%% file: main.tex
\documentclass[10pt,twocolumn,letterpaper]{article}

\usepackage[pagenumbers]{cvpr} 
\usepackage{multirow}
\usepackage{booktabs}
\usepackage{algorithm}
\usepackage{algorithmic}
\usepackage{xcolor}
\newcommand{\mname}{PointAlign} 
\usepackage{array}
\definecolor{cvprblue}{rgb}{0.21,0.49,0.74}
\usepackage[pagebackref,breaklinks,colorlinks,allcolors=cvprblue]{hyperref}

\def\paperID{2304} 
\def\confName{CVPR}
\def\confYear{2026}

\title{PointAlign: Feature-Level Alignment Regularization
for 3D Vision-Language Models}

\author{
Yuanhao Su$^{1, 2}$, Shaofeng Zhang$^{1}$\thanks{Shaofeng Zhang is the corresponding author}, Xiaosong Jia$^{3}$, Qi Fan$^{4}$ \\
$^{1}$University of Science and Technology of China, $^2$Fuzhou University, \\ $^{3}$Fudan University, $^{4}$Nanjing University \\
{\tt\small yharoldsu0627@gmail.com, sfzhang@ustc.edu.cn}
}

\begin{document}
\maketitle

\begin{abstract}
The development of 3D Vision-Language Models (VLMs), crucial for applications in robotics, autonomous driving, and augmented reality, is severely constrained by the scarcity of paired 3D-text data. Existing methods rely solely on next-token prediction loss, using only language tokens for supervision. This results in inefficient utilization of limited 3D data and leads to a significant degradation and loss of valuable geometric information in intermediate representations.
To address these limitations, we propose {\mname}, a novel feature-level alignment regularization method. {\mname} explicitly supervises intermediate point cloud tokens to preserve fine-grained 3D geometric-semantic information throughout the language modeling process.  
Specifically, we constrain the intermediate point cloud tokens within the LLM to align with visual input tokens via a consistency loss. By training only a lightweight alignment projector and LoRA adapters, {\mname} achieves explicit feature-level supervision with minimal computational overhead, effectively preventing geometric degradation.
Extensive experiments on ModelNet40 and Objaverse datasets demonstrate that our method achieves \textbf{2.08} pp improvement on average for classification tasks, with a substantial \textbf{7.50} pp gain on the challenging open-vocabulary Objaverse classification task and \textbf{4.88} pp improvement on 3D object captioning evaluated by Qwen2-72B-Instruct, validating the effectiveness of {\mname}. Code is publicly available at \href{https://github.com/yharoldsu0627/PointAlign}{https://github.com/yharoldsu0627/PointAlign}.
\end{abstract}

\section{Introduction}
\label{sec:intro}

While 2D vision-language models (VLMs)~\cite{li2023blip,dai2023instructblip} have achieved remarkable success through large-scale image-text datasets that enable effective cross-modal alignment~\cite{wang2022multi, su2025fame}, 3D vision-language models remain comparatively underdeveloped despite their importance for applications in robotics~\cite{zheng2024gaussiangrasper, zhang2025flowpolicy}, autonomous driving~\cite{li2022deepfusion, mao20233d}, and augmented reality~\cite{fan2022fully,chen2023voxelnext}. A fundamental challenge in 3D VLM development is the scarcity of high-quality 3D-text paired data—unlike easily accessible 2D images~\cite{schuhmann2022laion}, 3D point clouds~\cite{wu20153d} require expensive acquisition, resulting in smaller datasets with often simplistic textual descriptions~\cite{tang2025more}. This data limitation poses a core research question: \textbf{how to maximize knowledge extraction from scarce supervision for 3D understanding}.

Recent efforts such as PointLLM~\cite{xu2024pointllm}, ShapeLLM~\cite{qi2024shapellm}, and MiniGPT-3D~\cite{tang2024minigpt} have made progress in addressing these issues. PointLLM and ShapeLLM achieve competitive performance via full-model fine-tuning at substantial computational cost, while MiniGPT-3D reduces costs by leveraging 2D VLM priors. However, these methods optimize through language modeling objectives, which only reward geometric features that directly facilitate next-token prediction. This creates a critical issue: structural cues valuable for spatial reasoning but orthogonal to immediate language tasks may be discarded during training, a degradation we confirm in our feature quality analysis (see Figure~\ref{fig:knn}). Recent work on 2D vision-language models~\cite{kaduri2025s,yoon2025visual, li2025spatial} has shown that without explicit visual supervision, representational quality degrades across network depth. In the context of point cloud understanding, this problem is amplified by limited 3D training data. Point cloud networks process information hierarchically: initial layers establish object-level context, intermediate layers resolve part-level geometry, and final layers produce language outputs~\cite{yoon2025visual}. Therefore, preserving structural information throughout the forward pass is critical for accurate 3D reasoning.

To address this, we propose {\mname}, a novel feature-level alignment regularization method. Our method is built on a key insight: the representations from the initial stage already capture fine-grained geometric and semantic details, serving as an ideal internal supervision target. We directly align the LLM's intermediate point cloud tokens with these high-quality, initial-stage features. This alignment is enforced by a consistency loss during the second training stage. This approach introduces explicit geometric guidance with minimal overhead, as we only update a lightweight alignment projector (only 8.39M parameters) and LoRA adapters while freezing all pre-trained modules.

We conduct extensive experiments on several 3D vision-language benchmarks, including 3D classification and captioning. \textcolor{black}{Results show that our alignment regularization improves performance 
across nearly all evaluated tasks}, confirming that explicit feature-level supervision helps better preserve geometric structure (as validated by Figure~\ref{fig:knn}) and exploit limited 3D data (as demonstrated in Figure~\ref{fig:fraction}). On average,~\mname~improves classification accuracy by \textbf{2.08} pp on ModelNet40 and Objaverse. Notably, it achieves a striking \textbf{7.50} pp gain on the challenging open-vocabulary Objaverse classification task. For 3D object captioning on Objaverse, evaluated by Qwen2-72B-Instruct, our method outperforms the baseline by a significant \textbf{4.88} pp, demonstrating strong generalization in open-domain 3D vision-language scenarios of the proposed {\mname}.

\section{Related Work}
\label{sec:related}

\textbf{Large 3D Point Cloud-Language Models. }
\label{sec:related:3dllm}
The success of large language models in reasoning and generalization has motivated efforts to develop 3D point cloud-language understanding systems. Initial methods~\cite{xue2023ulip,xue2024ulip,zhang2022pointclip,zhu2023pointclip,huang2023clip2point,qi2023contrast,liu2023openshape,wang2023take,guo2023joint,dong2022autoencoders} adopt an indirect approach by converting 3D shapes into multiple 2D projections, then leveraging pre-trained vision-language models for cross-modal alignment~\cite{cr2pq, zhang2021zero, zhang2022align}. While computationally efficient, this rendering-based pipeline discards native geometric structure, hindering models from capturing spatial relationships inherent to 3D data. To address this limitation, recent work~\cite{xu2024pointllm,qi2024shapellm,qi2024gpt4point,guo2023point,tang2024minigpt} directly processes point cloud tokens through learned projection modules that bridge the gap between 3D encoders and language models. Full model fine-tuning approaches like PointLLM~\cite{xu2024pointllm} and ShapeLLM~\cite{qi2024shapellm} achieve strong performance but at significant computational cost. For example, PointLLM-13B demands over 200 GPU-hours on high-end accelerators. MiniGPT-3D~\cite{tang2024minigpt} mitigates this burden by incorporating 2D vision-language priors to facilitate 3D-text alignment, achieving comparable results with substantially reduced training resources. Beyond training efficiency, recent work has also explored inference acceleration for 3D MLLMs through visual token pruning techniques~\cite{huang2025fast3d}.

\textbf{Representation Supervision in Multimodal LLMs.}
\label{sec:related:infoflow}
Recent studies on 2D vision-language models reveal that visual embeddings can lose fidelity across network depth when supervision signals are weak~\cite{kaduri2025s,jiang2025devils,kang2025your}. To address this issue, reconstruction-based approaches~\cite{he2022masked,wang2025visual,wang2024reconstructive, pqae, pcpmae} have been proposed to supervise intermediate representations by recovering visual inputs, demonstrating effectiveness in retaining fine-grained details for 2D understanding tasks. However, 3D point cloud reasoning demands fundamentally different capabilities—models must capture structural relationships and geometric configurations that extend beyond low-level visual features~\cite{zhang2023tale,tong2024cambrian}. In the 3D point cloud domain, methods leveraging pre-trained 3D encoders~\cite{huang2025mllms, li2025spatial} provide specialized geometric supervision by distilling knowledge from foundation models trained on large-scale 3D point cloud data, though this introduces additional computational overhead during the training phase.

\section{The Proposed {\mname}}

\textbf{Preliminaries: MiniGPT-3D}

MiniGPT-3D~\cite{tang2024minigpt} is an efficient 3D large language model that aligns 3D point clouds with LLMs by leveraging priors from 2D vision-language models. Given a point cloud $\mathbf{P} \in \mathbb{R}^{n \times d}$, where $n$ denotes the number of points and $d$ represents the feature dimension of each point, the point cloud encoder $f_{pc}(\cdot)$ first encodes it into 3D features $f_{pc}(\mathbf{P}) \in \mathbb{R}^{m \times D}$, where $m$ is the length of the feature sequence and $D$ is the feature dimension. Subsequently, an MLP projection layer $f_{MLP}(\cdot)$ projects the point cloud features into the semantic space of Q-Former. Next, Q-Former~\cite{li2023blip} $f_{QF}(\cdot)$ queries the point cloud features through learnable query vectors $\mathbf{Q} \in \mathbb{R}^{o \times D_1}$, yielding point cloud tokens $\overline{\mathbf{Q}} = f_{QF}(f_{MLP}(f_{pc}(\mathbf{P})), \mathbf{Q}) \in \mathbb{R}^{o \times D_1}$, where $o$ is the number of queries and $D_1$ is the hidden dimension of Q-Former. Finally, the modality projector maps the Q-Former output to the input space of the LLM as $\mathbf{T}_{pc} \in \mathbb{R}^{o \times C}$, where $C$ is the shared dimension between point cloud and text tokens. The point cloud features are concatenated with text features $\mathbf{T} = [\mathbf{T}_{text}, \mathbf{T}_{pc}]$ and fed into the large language model $f_{llm}(\cdot)$ to autoregressively generate text responses. The first three recipes of MiniGPT-3D train the Q-Former, projector, and LLM using point cloud-text paired data, with the training objective being the cross-entropy loss for next token prediction $\mathcal{L}_{ntp} = \text{CrossEntropy}(h(G), \overline{\mathbf{T}})$, where $h(\cdot)$ is the LLM's tokenizer, $G$ is the ground truth answer text, and $\overline{\mathbf{T}}$ is the predicted probability distribution.

\subsection{Overall Architecture}

\begin{figure}[tb!]
\centering
\includegraphics[width=\linewidth]{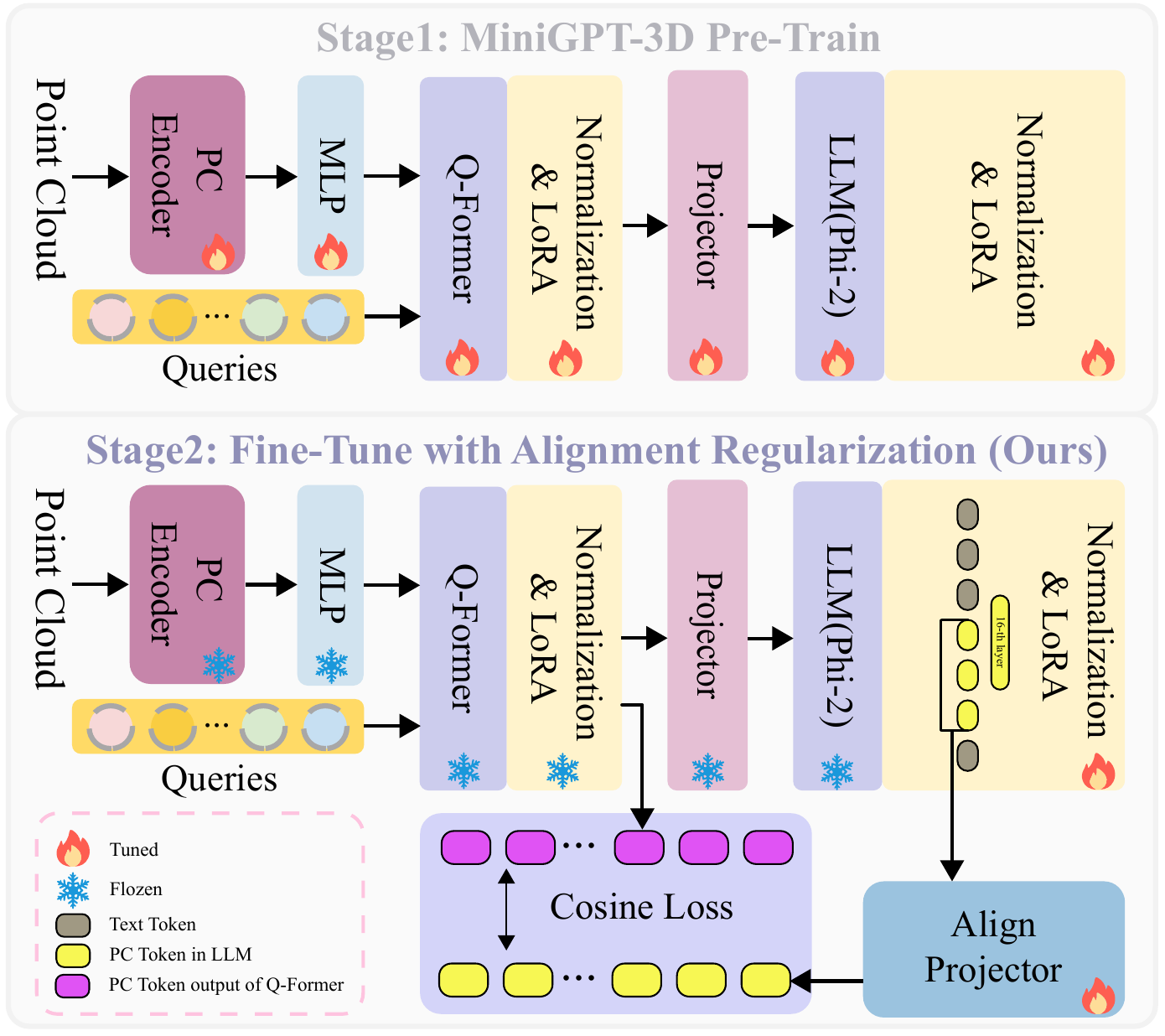}
\vspace{-20pt}
\caption{Overview of our method. Stage~1 adopts the three training recipes of MiniGPT-3D for pre-training. Stage~2 freezes the point cloud encoder, MLP, Q-Former, and modality projector, and only trains the LoRA layers of the LLM and the alignment projector. The alignment projector aligns the latent representations of point cloud tokens in the LLM with the Q-Former output through cosine similarity loss. Flame icons indicate trainable modules, and snowflakes indicate frozen modules.}
\label{fig:framework}
\end{figure}

As illustrated in Figure~\ref{fig:framework}, our method employs a two-stage training strategy to enhance the geometric understanding capabilities of MiniGPT-3D. Stage~1 is the MiniGPT-3D pre-training, which adopts the three training recipes of MiniGPT-3D and trains the point cloud encoder, MLP projection layer, Q-Former, modality projector, and LLM backbone. Stage~2 is the alignment regularization fine-tuning, which freezes the point cloud encoder, MLP projection layer, Q-Former, and modality projector, and only trains the LoRA layers~\cite{hu2022lora} of the LLM and the newly introduced alignment projector composed of 3~linear layers. \textcolor{black}{Note that the alignment projector is used only during training and is discarded 
at inference, introducing zero additional inference overhead.}

The core idea is to align the point cloud tokens in the intermediate layers of trainable LLM with the point cloud tokens output by frozen Q-Former. Specifically, the alignment projector maps the point cloud tokens $\mathbf{T}_{pc}^{(\ell)}$ at the $\ell$-th layer of the LLM to the Q-Former feature space, constraining the representations to remain consistent with the Q-Former output $\overline{\mathbf{Q}}$ through cosine similarity loss, thereby preserving richer point cloud geometric and semantic information during the language modeling process.

\begin{algorithm}[tb!]
\caption{Fine-tune with Alignment Regularization.}
\label{alg:stage2}
\begin{algorithmic}[1]
\REQUIRE Point cloud $\mathbf{P}$, text prompt features $\mathbf{T}_{text}$, ground truth $G$, pre-trained components from Stage~1, \textcolor{black}{hyperparameters $\lambda$ and $\ell$}
\ENSURE Fine-tuned LLM with LoRA and alignment projector $f_{\pi}$

\STATE Freeze: $f_{pc}, f_{MLP}, f_{QF}, f_{proj}$
\STATE Initialize LoRA in $f_{llm}$ and 3-layer projector $f_{\pi}$ \COMMENT{Eq.~(1)}

\FOR{each training batch}
    \STATE $\overline{\mathbf{Q}} \leftarrow f_{QF}(f_{MLP}(f_{pc}(\mathbf{P})), \mathbf{Q})$ \COMMENT{Q-Former output}
    \STATE $\mathbf{T}_{pc} \leftarrow f_{proj}(\overline{\mathbf{Q}})$ \COMMENT{Modality projection}
    \STATE $\mathbf{T} \leftarrow [\text{BOS}, \mathbf{T}_{text}, \mathbf{T}_{pc}, h(G)]$
    
    \STATE $s \leftarrow 1 + |\mathbf{T}_{text}|, \quad e \leftarrow s + 32$ \COMMENT{PC token positions}
    \STATE $\overline{\mathbf{T}}, \{H^{(1)}, \ldots, H^{(L)}\} \leftarrow f_{llm}(\mathbf{T})$ \COMMENT{Output \& hidden states}
    
    \STATE $\mathbf{T}_{pc}^{(\ell)} \leftarrow H^{(\ell)}[:, s:e, :]$ \COMMENT{Extract PC tokens from Layer $\ell$}
    \STATE $\tilde{\mathbf{Q}} \leftarrow f_{\pi}(\mathbf{T}_{pc}^{(\ell)})$ \COMMENT{Project to Q-Former space}
    
    \STATE $\mathcal{L}_{ntp} \leftarrow \text{CrossEntropy}(h(G),\overline{\mathbf{T}})$
    \STATE $\mathcal{L}_{align}$ by Eq.~(2) with $\overline{\mathbf{Q}}^{\text{detach}}$
    \STATE $\mathcal{L}_{total}$ by Eq.~(3)
    
    \STATE Update $\theta_{LoRA}$ and $\theta_{\pi}$ via $\nabla \mathcal{L}_{total}$
\ENDFOR
\end{algorithmic}
\end{algorithm}

\begin{table*}[tb!]
\centering
\renewcommand{\arraystretch}{1.3} 
\caption{Performance evaluation of generative 3D object classification on ModelNet40 test set and Objaverse dataset. Classification accuracy (\%) is evaluated using two prompt formats: Instruction-based (I) ``What is this?'' and Completion-based (C) ``This is an object of'', with average scores provided. Best and second-best performances are highlighted in bold and underlined, respectively.}
\vspace{-8pt}
\label{tab:classification}
\resizebox{\textwidth}{!}{
\begin{tabular}{lccccccccccc}
\toprule
\multirow{2}{*}[-2pt]{Model} & \multirow{2}{*}[-2pt]{Reference} & 
\multirow{2}{*}[-2pt]{\begin{tabular}{@{}c@{}}LLM \\ Size\end{tabular}} & 
\multirow{2}{*}[-2pt]{\begin{tabular}{@{}c@{}}3D Data \\ Size\end{tabular}} & 
\multirow{2}{*}[-2pt]{Input} & 
\multicolumn{3}{c}{ModelNet40} & \multicolumn{3}{c}{Objaverse} & 
\multirow{2}{*}[-2pt]{Average} \\
\cmidrule(lr){6-8} \cmidrule(lr){9-11}

& & & & & (I) & (C) & Average & (I) & (C) & Average & \\
\midrule
InstructBLIP-7B~\cite{dai2023instructblip} & NeurIPS,23 & 7B & 0K & Single-V. Img. & 17.67 & 22.81 & 20.24 & 21.50 & 26.00 & 23.75 & 22.00 \\
InstructBLIP-13B~\cite{dai2023instructblip} & NeurIPS,23 &   13B & 0K & Single-V. Img. & 21.56 & 21.92 & 21.74 & 21.50 & 21.50 & 21.50 & 21.62 \\
LLaVA-7B~\cite{liu2023visual} & CVPR,24 & 7B & 0K & Single-V. Img. & 27.11 & 21.68 & 24.40 & 37.50 & 30.00 & 33.75 & 29.07 \\
LLaVA-13B~\cite{liu2023visual} & CVPR,24 & 13B & 0K & Single-V. Img. & 27.71 & 27.76 & 27.74 & 39.50 & 35.50 & 37.50 & 32.62 \\
GPT-4o mini~\cite{menick2024gpt} & OpenAI & - & 0K & Single-V. Img. & 22.00 & 23.10 & 22.55 & 39.00 & 35.00 & 37.00 & 29.78 \\
\midrule
Point-Bind LLM~\cite{guo2023point} & arXiv,23 & 7B & - & Point Cloud & 46.60 & 45.02 & 45.81 & 7.50 & 7.58 & 7.54 & 26.68 \\
GPT4Point~\cite{qi2024gpt4point} & CVPR,24 & 2.7B & 660K & Point Cloud & 21.39 & 21.07 & 21.23 & 49.00 & 46.50 & 47.75 & 34.49 \\
PointLLM-7B~\cite{xu2024pointllm} & ECCV,24 & 7B & 730K & Point Cloud & 51.34 & 50.36 & 50.85 & 62.00 & 63.00 & 62.50 & 56.68 \\
PointLLM-13B~\cite{xu2024pointllm} & ECCV,24 & 13B & 730K & Point Cloud & 51.70 & 52.67 & 52.19 & 61.50 & 63.00 & 62.25 & 57.22 \\
MiniGPT-3D~\cite{tang2024minigpt} & MM,24 & 2.7B & 730K & Point Cloud & \textbf{61.99} & \underline{60.49} & \textbf{61.24} & \underline{65.00} & \underline{68.50} & \underline{66.75} & \underline{64.00} \\

\midrule
\textbf{Ours} & - & 2.7B & 730K & Point Cloud & \underline{61.55} & \textbf{60.78} & \underline{61.17} & \textbf{72.50} & \textbf{69.50} & \textbf{71.00} & \textbf{66.08} \\
\bottomrule
\end{tabular}
}
\end{table*}

\subsection{Point Cloud Representation Alignment}
The training of MiniGPT-3D and other 3D understanding methods~\cite{xu2024pointllm} is commonly optimized solely through $\mathcal{L}_{ntp}$, leading to point cloud features only receiving indirect supervision from text prediction. Let $\mathbf{T}_{pc}^{(\ell)} \in \mathbb{R}^{o \times C}$ denote the point cloud tokens at the $\ell$-th layer of the LLM (The specific choice of $\ell$ is detailed in the experimental settings). During training, no explicit regularization is applied to the $\mathbf{T}_{pc}^{(\ell)}$, potentially losing fine-grained 3D geometric information. To address this, we propose to align the point cloud tokens in the LLM's intermediate layers with the Q-Former output.

We select the Q-Former output $\overline{\mathbf{Q}} \in \mathbb{R}^{o \times D_1}$ \textcolor{black}{(features after Q-Former and its subsequent Normalization and LoRA layers)} as the alignment target rather than the point cloud encoder output or deep LLM representations. The rationale behind this choice is that the Q-Former has learned the mapping between geometry and semantics through point cloud-text pairs during Stage~1 training~\cite{li2023blip,tang2024minigpt}, and its output contains both geometric and semantic information. Compared to the point cloud encoder that only captures geometric features, the Q-Former output is more suitable for supervising the LLM's intermediate layers. Meanwhile, $\overline{\mathbf{Q}}$ is primarily supervised by point cloud-text pairs directly before entering the LLM, preserving more original geometric details, while deep LLM representations may have lost 3D information through multiple layers of language modeling.




\textcolor{black}{We further propose an alignment projector $f_{\pi}: \mathbb{R}^{C} \to \mathbb{R}^{D_1}$, which maps the point cloud tokens at the $\ell$-th layer of the LLM to the feature space of Q-Former.} As shown in Figure~\ref{fig:framework}, the projector consists of three linear layers and two SiLU~\cite{elfwing2018sigmoid} activation functions, providing sufficient nonlinear transformation capacity while maintaining computational efficiency. Let $\mathbf{W}_i$ and $\mathbf{b}_i$ denote the weight matrix and bias vector of the $i$-th linear transformation. The mapping process is formulated as:
\begin{equation}
\begin{aligned}
\mathbf{h}^{(1)} &= \mathbf{W}_1 \cdot \mathbf{T}_{pc}^{(\ell)} + \mathbf{b}_1, \\
\mathbf{h}^{(2)} &= \mathbf{W}_2 \cdot \text{SiLU}(\mathbf{h}^{(1)}) + \mathbf{b}_2, \\
\tilde{\mathbf{Q}} &= \mathbf{W}_3 \cdot \text{SiLU}(\mathbf{h}^{(2)}) + \mathbf{b}_3,
\end{aligned}
\end{equation}
where $\mathbf{W}_1 \in \mathbb{R}^{d_h \times C}$, $\mathbf{W}_2 \in \mathbb{R}^{d_h \times d_h}$, $\mathbf{W}_3 \in \mathbb{R}^{D_1 \times d_h}$, and $d_h$ is the hidden dimension. The SiLU activation function is defined as $\text{SiLU}(x) = x \cdot \sigma(x)$, where $\sigma(x)$ is the sigmoid function. The final output is $\tilde{\mathbf{Q}} \in \mathbb{R}^{o \times D_1}$.

The alignment loss is defined using cosine similarity~\cite{7780803}, as shown in Figure~\ref{fig:framework}. We adopt the cosine loss as an alignment loss because it focuses on feature direction rather than magnitude, making it well-suited for aligning representations across different feature spaces. Given the mapped features $\tilde{\mathbf{Q}}$ and the Q-Former output $\overline{\mathbf{Q}}$, the alignment loss is:
\begin{equation}
\mathcal{L}_{align} = -\frac{1}{o}\sum_{i=1}^{o} \frac{\tilde{\mathbf{Q}}_i^\top \overline{\mathbf{Q}}_i}{\|\tilde{\mathbf{Q}}_i\|_2 \|\overline{\mathbf{Q}}_i\|_2}, 
\end{equation}
where $i$ denotes the $i$-th query token.

\begin{table*}[tb!]
\centering
\renewcommand{\arraystretch}{1.3} 
\caption{Performance evaluation of 3D object captioning on the Objaverse dataset. Evaluation metrics include Qwen2 evaluation and traditional metrics. Bold and underlined values represent the best and second-best performances, respectively.}
\vspace{-8pt}
\label{tab:captioning}
\resizebox{\textwidth}{!}{

\begin{tabular}{lcccccccc}
\toprule
Model & Reference & \begin{tabular}{@{}c@{}}LLM \\ Size\end{tabular} & \begin{tabular}{@{}c@{}}3D Data \\ Size\end{tabular} & Input & Qwen2-72B-Instruct & Sentence-BERT & SimCSE & Average \\
\midrule
InstructBLIP-7B~\cite{dai2023instructblip} & NIPS,23 & 7B & 0K & Single-V. Img. & 16.10 & 35.79 & 36.67 & 29.52 \\
InstructBLIP-13B~\cite{dai2023instructblip} & NIPS,23 & 13B & 0K & Single-V. Img. & 13.79 & 33.52 & 35.60 & 27.64 \\
LLaVA-1.5-7B~\cite{liu2023visual} & CVPR,24 & 7B & 0K & Single-V. Img. & 17.80 & 39.32 & 41.08 & 32.73 \\
LLaVA-1.5-13B~\cite{liu2023visual} & CVPR,24 & 13B & 0K & Single-V. Img. & 16.00 & 39.64 & 40.90 & 32.18 \\
GPT-4o mini~\cite{menick2024gpt} & OpenAI & - & 0K & Single-V. Img. & 26.00 & 38.70 & 39.13 & 34.61 \\
\midrule
Point-Bind LLM~\cite{guo2023point} & arXiv,23 & 7B & - & Point Cloud & 1.93 & 27.29 & 25.35 & 18.19 \\
GPT4Point~\cite{qi2024gpt4point} & CVPR,24 & 2.7B & 660K & Point Cloud & 21.75 & 41.10 & 41.24 & 34.70 \\
PointLLM-7B~\cite{xu2024pointllm} & ECCV,24 & 7B & 730K & Point Cloud & 42.20 & 48.50 & 48.92 & 46.54 \\
PointLLM-13B~\cite{xu2024pointllm} & ECCV,24 & 13B & 730K & Point Cloud & 40.40 & 49.07 & 48.41 & 45.96 \\
MiniGPT-3D~\cite{tang2024minigpt} & MM,24 & 2.7B & 730K & Point Cloud & \underline{48.17} & \underline{49.54} & \textbf{51.39} & \underline{49.70} \\
\midrule
\textbf{Ours} & - & 2.7B & 730K & Point Cloud & \textbf{53.05} & \textbf{49.94} & \underline{51.32} & \textbf{51.44} \\
\bottomrule
\end{tabular}
}
\end{table*}

\begin{figure*}[tb!]
\centering
\includegraphics[width=\textwidth]{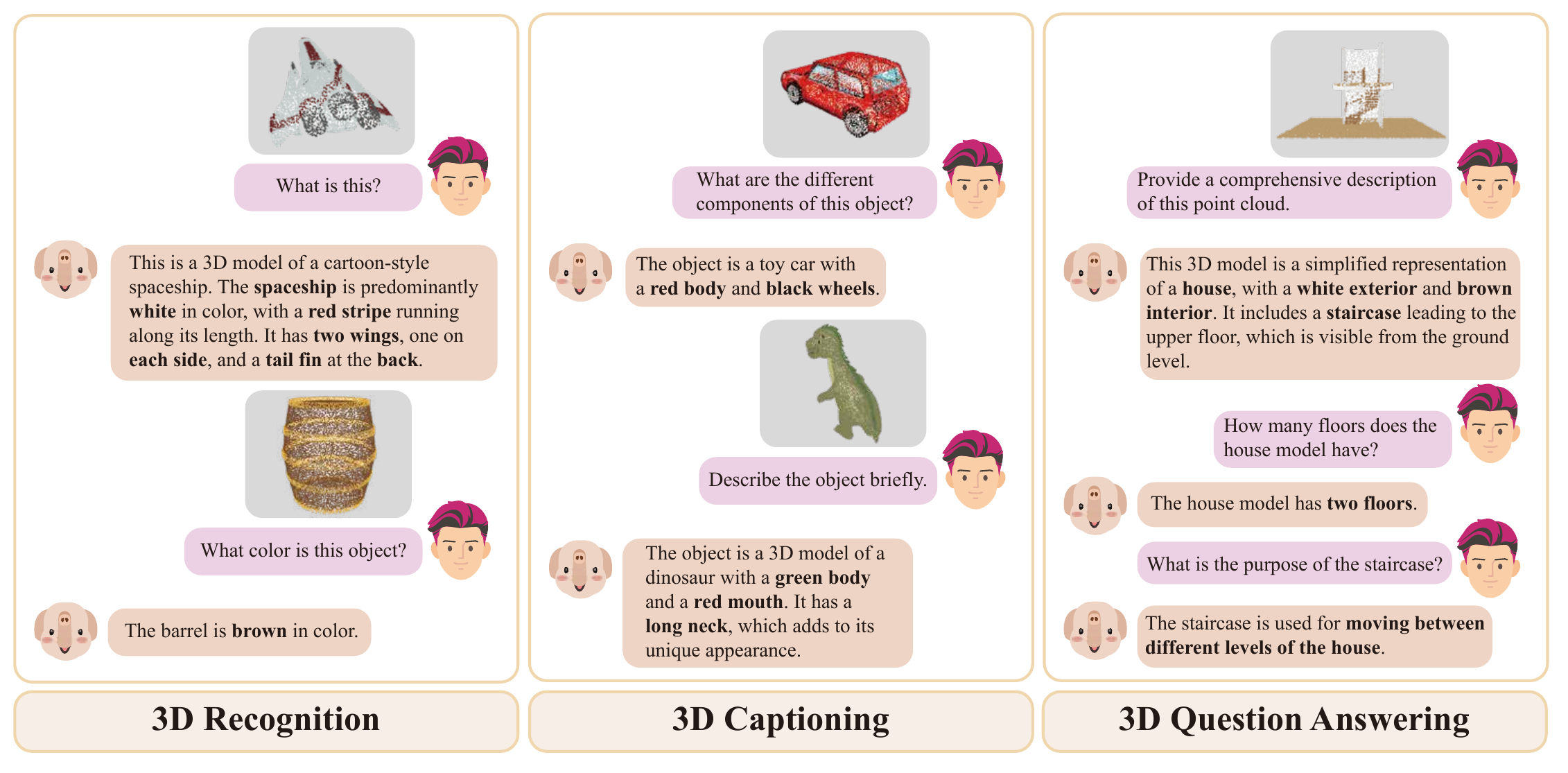}
\vspace{-20pt}
\caption{Examples of 3D Object Understanding with Our Model. The figure demonstrates our model's capabilities in the context of 3D object understanding, showcasing its performance on tasks such as 3D object recognition, description generation, and 3D VQA.}
\label{fig:qualitative_examples}
\end{figure*}

\begin{table}[tb!]
\centering
\caption{Qualitative comparison of model outputs for classification and captioning on the Objaverse datasets.}
\vspace{-8pt}
\label{tab:Qualitative}
\includegraphics[width=\linewidth]{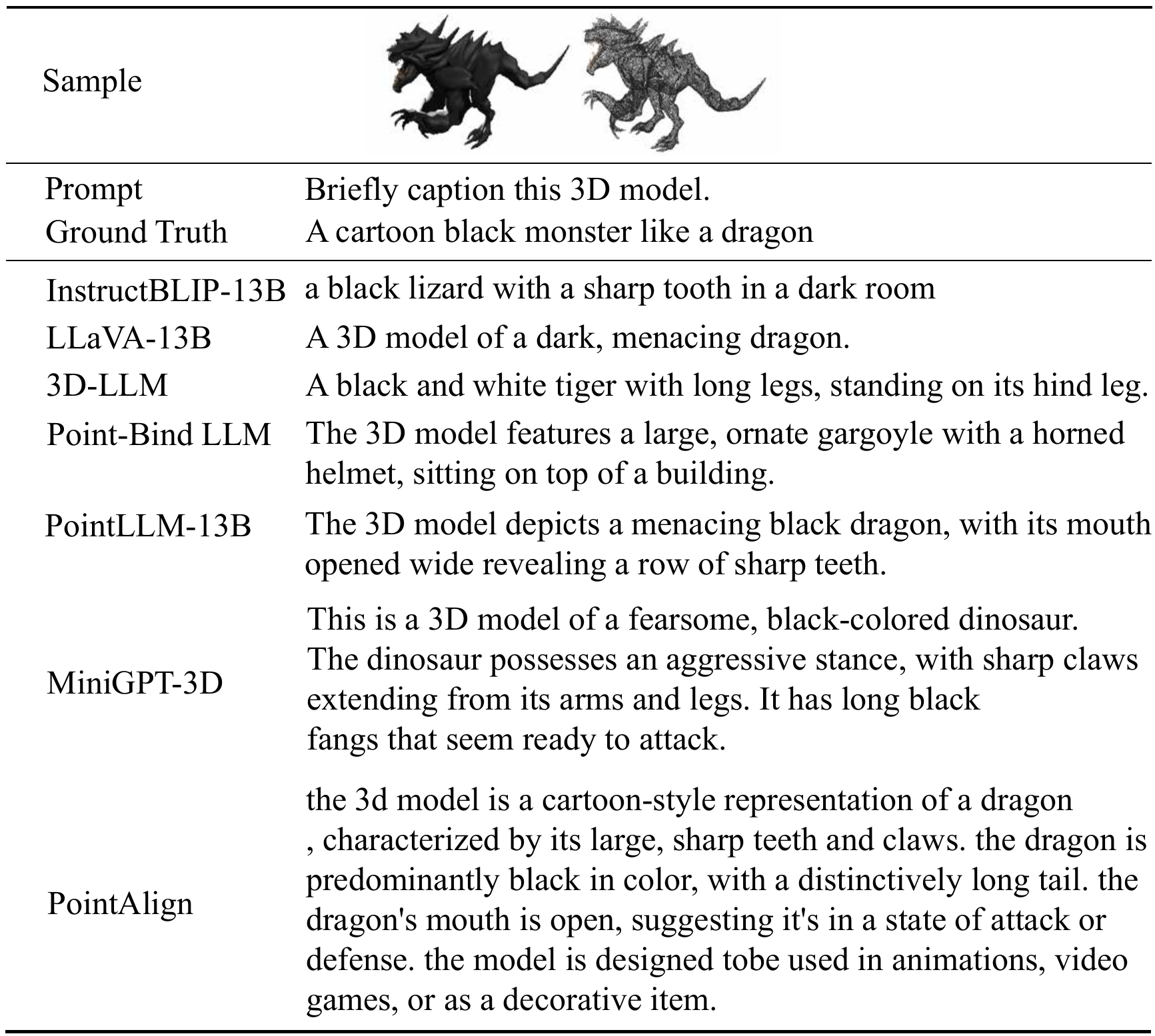}
\end{table}

\subsection{Training Objective}

The total loss function consists of the next token prediction loss and the alignment loss:
\begin{equation}
\mathcal{L}_{total} = \mathcal{L}_{ntp} + \lambda \mathcal{L}_{align},
\end{equation}
where $\lambda$ is a hyperparameter balancing the two losses. During Stage~2, we freeze the parameters of the point cloud encoder, MLP projection layer, Q-Former, and modality projector, and only train the LoRA layers of the LLM and the alignment projector $f_{\pi}$. The Q-Former output $\overline{\mathbf{Q}}$ is treated as a constant when computing the alignment loss and does not participate in gradient backpropagation, preserving the point cloud-language alignment knowledge learned in Stage~1. This training strategy preserves pre-trained knowledge while efficiently fine-tuning the LLM through LoRA and preserving point cloud geometric and semantic information through alignment constraints.

\subsection{Algorithm Pseudo Code}

Algorithm~\ref{alg:stage2} outlines the complete training procedure of our alignment regularization approach. As shown in the pseudo-code, the implementation is straightforward yet effective: by freezing pre-trained modules and introducing a lightweight alignment projector (3~linear layers), followed by an alignment loss, we achieve significant improvements in geometric understanding with minimal computational overhead and easy integration into existing pipelines.

\section{Experiments}

\subsection{Experimental Settings}
We employ Phi-2~\cite{javaheripi2023phi} as the foundational component for the large language model and utilize PointBERT~\cite{yu2022point} to extract semantic features from 3D point clouds. All experiments are conducted on NVIDIA RTX 4090 GPUs equipped with 24GB of memory. We adopt a two-stage training strategy. In the pre-training stage, we follow the first three training stages of MiniGPT-3D~\cite{tang2024minigpt}, using the same data scale, training epochs, and learning rate schedule. In the fine-tuning stage, we train for 10000 iterations on 70k detailed descriptions and dialogue data, with the learning rate linearly decaying from 3e-6 to 1e-6. Meanwhile, we introduce an alignment module consisting of three linear layers with SiLU~\cite{elfwing2018sigmoid} activation functions to achieve feature dimensionality reduction and alignment. \textcolor{black}{The alignment is applied at the $16$-th layer of the Phi-2 
model.} The training data is sourced from the point-text instruction dataset~\cite{xu2024pointllm}, encompassing both brief description instructions and complex instruction samples. Following the data split criteria established by PointLLM~\cite{xu2024pointllm}, we reserve 200 object samples for testing and evaluation. Regarding the point cloud input format, each sample comprises $n = 8192$ sampled points, with each point containing $d = 6$ dimensional features that encode both 3D spatial coordinates and corresponding attributes. To validate the effectiveness of our proposed method, we select several state-of-the-art models as baseline comparisons, specifically including 5 open-source 3D models~\cite{guo2023point,qi2024gpt4point,xu2024pointllm,tang2024minigpt} and 5 open-source 2D models~\cite{dai2023instructblip,liu2023visual,menick2024gpt}. Following GreenGLM~\cite{tang2025more}, we uniformly employ the Qwen2-72B-Instruct~\cite{team2024qwen2} model to assess and score the generation quality of all comparative methods.

\begin{table}[tb!]
\centering
\caption{Ablation study on loss functions. We evaluate the impact of different loss functions at 16-th layer on classification accuracy.}
\vspace{-8pt}
\label{tab:ablation_loss}
\begin{tabular}{lccccc}
\toprule
\multirow{2}{*}{Loss} & \multicolumn{2}{c}{ModelNet40} & \multicolumn{2}{c}{Objaverse} & \multirow{2}{*}{Average} \\
\cmidrule(lr){2-3} \cmidrule(lr){4-5}
& (I) & (C) & (I) & (C) & \\
\midrule
l1 & 61.30 & 60.58 & 71.50 & \textbf{69.50} & 65.72 \\
l2 & 61.51 & \textbf{61.75} & 69.50 & \textbf{69.50} & 65.57 \\
cosine & \textbf{61.55} & 60.78 & \textbf{72.50} & \textbf{69.50} & \textbf{66.08} \\
\bottomrule
\end{tabular}
\end{table}

\begin{table}[tb!]
\centering
\caption{Ablation study on target layers. We evaluate the impact of different layer indices on classification accuracy (\%).}
\vspace{-8pt}
\label{tab:ablation_layer}
\resizebox{\linewidth}{!}{
\begin{tabular}{lccccc}
\toprule
\multirow{2}{*}{Layer Index} & \multicolumn{2}{c}{ModelNet40} & \multicolumn{2}{c}{Objaverse} & \multirow{2}{*}{Average} \\
\cmidrule(lr){2-3} \cmidrule(lr){4-5}
& (I) & (C) & (I) & (C) & \\
\midrule
8 & 61.47 & 60.86 & 69.50 & \textbf{69.50} & 65.33 \\
12 & \textbf{61.71} & 60.33 & 70.50 & 68.00 & 65.14 \\
16 & 61.55 & 60.78 & \textbf{72.50} & \textbf{69.50} & \textbf{66.08} \\
20 & 61.30 & 60.09 & 70.50 & 69.00 & 65.22 \\
24 & \textbf{61.71} & 60.86 & 71.00 & 66.00 & 64.89 \\
28 & 61.30 & \textbf{60.94} & 70.00 & 67.00 & 64.81 \\
32 & 61.67 & 60.58 & 70.00 & 68.00 & 65.06 \\
12, 16, 20 & 61.63 & 60.21 & 69.00 & 66.50 & 64.34 \\
\bottomrule
\end{tabular}}
\end{table}

\begin{table}[tb!]
\centering
\caption{Ablation study on the weight of the alignment loss $\lambda$. We evaluate the impact of different $\lambda$ at 16-th layer.}
\vspace{-8pt}
\label{tab:ablation_lambda}
\begin{tabular}{lccccc}
\toprule
\multirow{2}{*}{$\lambda$} & \multicolumn{2}{c}{ModelNet40} & \multicolumn{2}{c}{Objaverse} & \multirow{2}{*}{Average} \\
\cmidrule(lr){2-3} \cmidrule(lr){4-5}
& (I) & (C) & (I) & (C) & \\
\midrule
0 & 60.58 & 60.29 & 69.50 & 68.50 & 64.72 \\
0.1 & 61.55 & 60.78 & \textbf{72.50} & \textbf{69.50} & \textbf{66.08} \\
0.3 & \textbf{61.75} & 60.21 & 69.50 & 67.50 & 64.74 \\
0.5 & 61.39 & \textbf{60.94} & 69.50 & 68.50 & 65.08 \\
0.7 & 61.10 & \textbf{60.94} & 67.00 & 68.50 & 64.39 \\
0.9 & 61.51 & 60.66 & 68.50 & 68.00 & 64.67 \\
\bottomrule
\end{tabular}
\end{table}

\subsection{Generative 3D Object Classification}
We conduct generative 3D object classification tasks on ModelNet40~\cite{wu20153d} and the more challenging open-vocabulary Objaverse~\cite{deitke2023objaverse} dataset. As shown in Table~\ref{tab:classification}, following~\cite{xu2024pointllm,qi2024shapellm}, we adopt two prompt formats for evaluation: instruction-based (I) ``What is this?'' and completion-based (C) ``This is an object of''. For zero-shot classification on the closed-set ModelNet40 dataset, we leverage the Qwen2-72B-Instruct model to parse the textual outputs from our method, enabling precise prediction from 40 candidate categories. For the open-vocabulary category recognition task on the Objaverse dataset, we employ Qwen2-72B-Instruct model as semantic judges to verify whether our model's generated descriptive text refers to the same object category as the ground truth label.

Experimental results demonstrate that our model achieves an average classification accuracy of 66.08\%, significantly outperforming all baseline methods. Compared to the current state-of-the-art 3D point cloud model MiniGPT-3D, we achieve a 2.08 pp improvement. More notably, on the Objaverse dataset under the instruction-based prompt, we obtain a substantial improvement of 7.50 pp over MiniGPT-3D. Additionally, the average accuracy reaches 71.00\%, representing a 4.25 pp improvement over MiniGPT-3D and a remarkable 8.75 pp gain over PointLLM-13B. This advantage is particularly critical in open-vocabulary scenarios, indicating that our model exhibits stronger generalization capabilities when handling open-domain categories. These results validate the effectiveness of aligning point cloud tokens in the intermediate layers of the LLM with the Q-Former output.

\vspace{-4pt}
\subsection{3D Object Captioning}
\vspace{-4pt}
The 3D object captioning task requires models to generate accurate and detailed natural language descriptions for input point clouds. We evaluate on the Objaverse dataset using the unified prompt ``Caption this 3D model in detail'' and employ Qwen2-72B-Instruct, Sentence-BERT, and SimCSE to assess the semantic similarity between model outputs and manual annotations. As shown in Table~\ref{tab:captioning}, our model achieves an average score of 51.44, surpassing all baseline methods. In the Qwen2 evaluation, we attain the highest score of 53.05, representing a 4.88 pp improvement over the second-best model MiniGPT-3D and a 10.85 pp gain over PointLLM-7B, while significantly outperforming 2D models such as GPT-4o mini. On the Sentence-BERT and SimCSE metrics, we achieve scores of 49.94 and 51.32, respectively, maintaining comparable performance to MiniGPT-3D. The experimental results demonstrate that our proposed partial token alignment mechanism effectively integrates 3D visual features with language semantic representations, enhancing the perception of fine-grained 3D object details while maintaining semantic accuracy.

\subsection{Qualitative Results}
Figure~\ref{fig:qualitative_examples} shows our model's performance on 3D recognition, captioning, and 3D question answering tasks. The examples demonstrate that our model generates accurate descriptions with details about object geometry, materials, and attributes. In the QA examples, the model correctly answers questions that require both visual understanding and external knowledge. The bolded text indicates accurate predictions that capture specific characteristics of the 3D objects.

\begin{figure}[tb!]
\centering
\includegraphics[width=\linewidth]{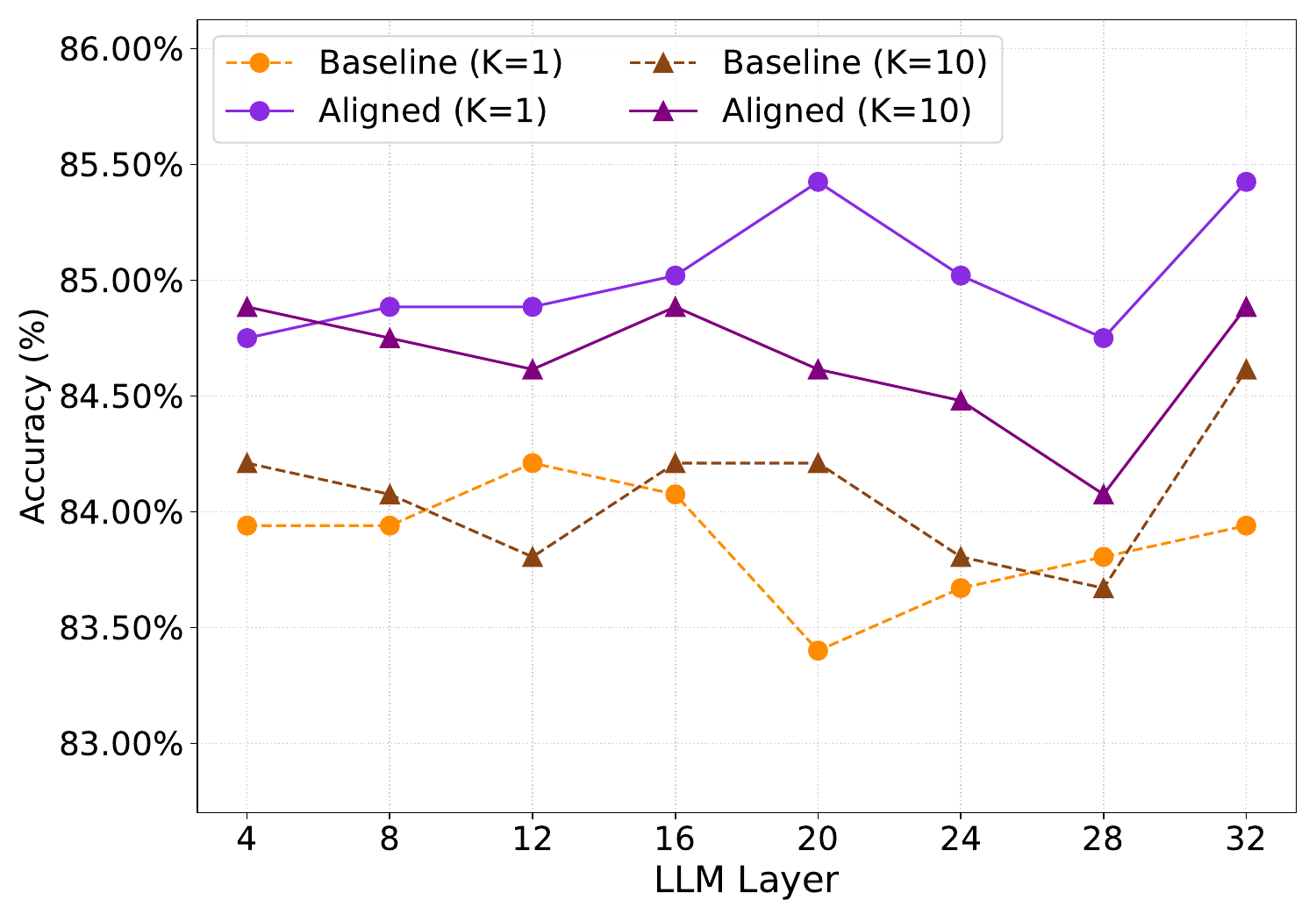}
\vspace{-20pt}
\caption{KNN classification accuracy of point cloud tokens extracted from different LLM layers on ModelNet40. We compare the baseline model and our aligned model using K=1 and K=10.}
\label{fig:knn}
\end{figure}

Table~\ref{tab:Qualitative} presents a qualitative comparison of outputs from different models on a representative sample. In this sample, 2D models and early 3D methods exhibit significant recognition deviations. Our model not only accurately determines the cartoon style of the dinosaur but also provides detailed descriptions of anatomical features such as sharp teeth and claws, recognizes the open mouth posture, and infers its attacking or defensive state along with potential application domains. These improvements stem from the partial token alignment mechanism that enforces consistency between intermediate layer representations and Q-Former outputs during training. This enables the model to retain richer point cloud semantic features rather than discarding valuable information due to insufficient textual supervision.

\subsection{Ablation Study}
In this section, ablation studies are first conducted to validate the key design choices of the alignment regularization method. Additional analyses are then performed to examine the quality of learned representations and data efficiency.

\textbf{Loss function.} We compare three distance metrics at layer 16, including L1, L2, and cosine similarity. As shown in Table~\ref{tab:ablation_loss}, cosine similarity achieves the best average accuracy of 66.08\%, outperforming L1 by 0.36 pp and L2 by 0.51 pp. Cosine loss focuses on directional consistency rather than numerical differences, making it more suitable for LLM hidden representations where point cloud tokens undergo multiple transformations while maintaining semantic consistency with Q-Former outputs.

\textbf{Target layer.} We evaluate alignment at layers 8, 12, 16, 20, 24, 28, and 32. Additionally, we test joint alignment at multiple layers 12, 16, and 20 simultaneously. As shown in Table~\ref{tab:ablation_layer}, layer 16 achieves the strongest performance with an average accuracy of 66.08\%, showing 0.75 pp improvement over layer 8 and 0.94 pp improvement over layer 12. Deeper layers from 20 to 32 exhibit gradual performance degradation, with layer 28 dropping 1.27 pp compared to layer 16. This indicates that intermediate layers are most effective for processing 3D visual information. Joint alignment at layers 12, 16, and 20 decreases performance by 1.74 pp, suggesting that simultaneous constraints at multiple layers introduce conflicting optimization objectives.

\textbf{Alignment weight.} \textcolor{black}{We investigate the impact of the alignment loss weight $\lambda$ at layer 16 
using cosine similarity. As shown in Table~\ref{tab:ablation_lambda}, 
$\lambda=0$ (NTP loss only, without alignment) yields 64.72\%, while 
$\lambda=0.1$ achieves the best average accuracy of 66.08\%, confirming that the gains stem from alignment regularization rather than merely Stage-2 training.} Higher weights from 0.3 to 0.9 lead to consistent performance drops, with $\lambda=0.7$ decreasing accuracy by 1.69 pp. This suggests that excessive alignment constraints overly restrict the model's capacity to fuse point cloud features with linguistic representations, while a smaller weight preserves sufficient flexibility for cross-modal feature integration. This flexibility appears crucial, as performance on the instruction-based Objaverse task drops by 3.0 pp as soon as $\lambda$ increases from 0.1 to 0.3.

\begin{figure}[tb!]
\centering
\includegraphics[width=\linewidth]{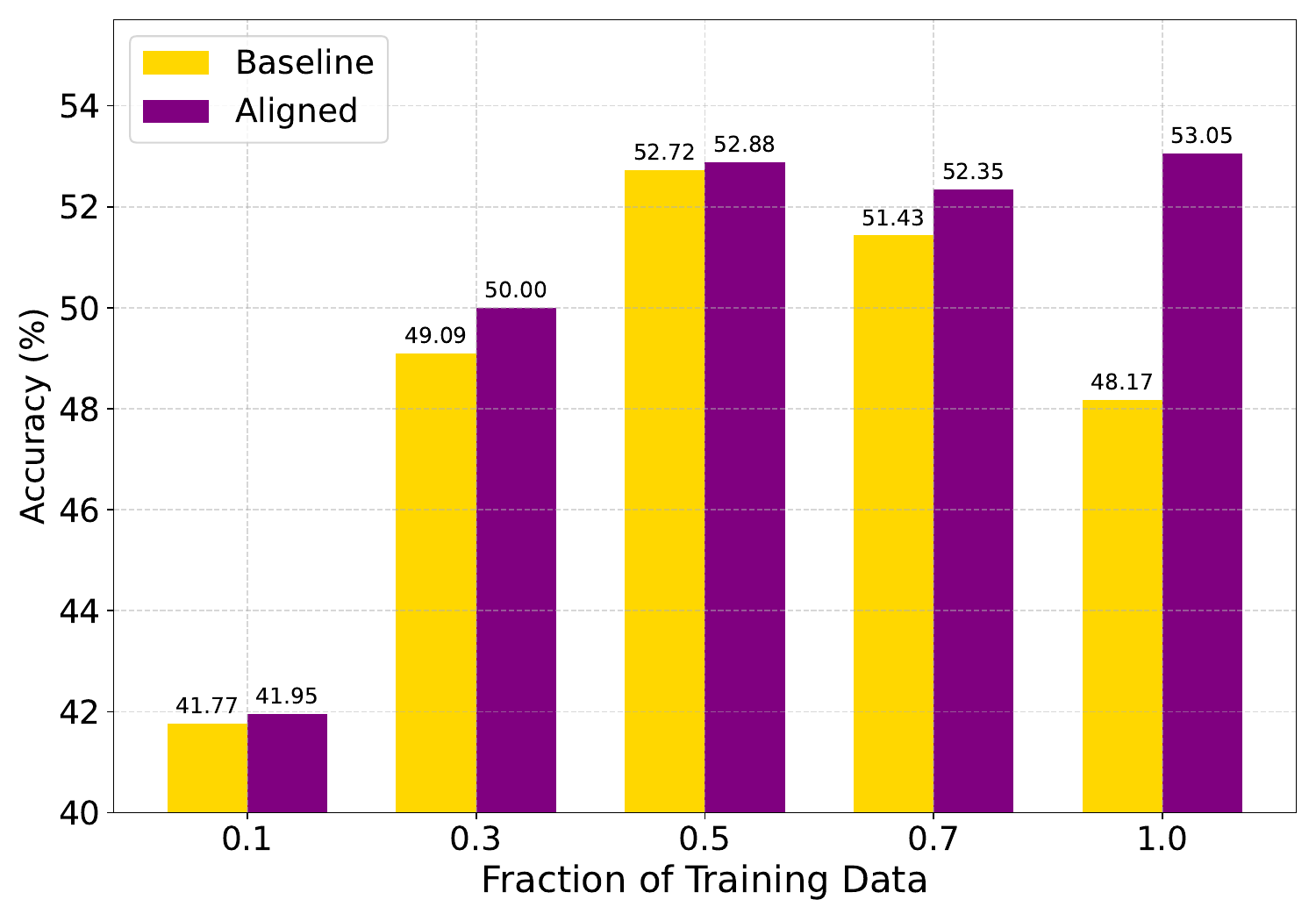}
\vspace{-20pt}
\caption{Impact of training data fraction on 3D object captioning performance. 
We evaluate baseline and aligned models using 10\%, 30\%, 50\%, 70\%, and 
100\% of training data on Objaverse dataset.}
\label{fig:fraction}
\end{figure}

\textbf{Number of linear layers.} We examine the impact of the projector architecture at layer 16 by varying the number of linear layers from 1 to 4. As shown in Table~\ref{tab:ablation_linear}, 3~linear layers achieve optimal performance with an average accuracy of 66.08\%, improving 1.63 pp over a single layer and 0.60 pp over two layers. The 4~layer configuration shows slight degradation to 65.24\%, indicating that deeper projectors risk overfitting. This accuracy of 65.24\% is notably lower than even the 2~layer model's performance, which strongly suggests the added complexity is detrimental. The 3~layer architecture provides adequate modeling capacity to bridge the representation gap between Q-Former and LLM hidden states without introducing unnecessary complexity.

\begin{table}[tb!]
\centering
\caption{Ablation study on the number of linear layers in the alignment projector. We evaluate the impact at layer 16 on classification accuracy using cosine similarity loss with $\lambda=0.1$.}
\vspace{-8pt}
\label{tab:ablation_linear}
\resizebox{\linewidth}{!}{
\begin{tabular}{lccccc}
\toprule
\multirow{2}{*}{Num. Layers} & \multicolumn{2}{c}{ModelNet40} & \multicolumn{2}{c}{Objaverse} & \multirow{2}{*}{Average} \\
\cmidrule(lr){2-3} \cmidrule(lr){4-5}
& (I) & (C) & (I) & (C) & \\
\midrule
1 & 61.47 & 60.82 & 70.50 & 65.00 & 64.45 \\
2 & 61.26 & \textbf{61.67} & 71.00 & 68.00 & 65.48 \\
3 & \textbf{61.55} & 60.78 & \textbf{72.50} & \textbf{69.50} & \textbf{66.08} \\
4 & 61.43 & 61.02 & 70.00 & 68.50 & 65.24 \\
\bottomrule
\end{tabular}}
\end{table}

\textbf{Alignment target.} We compare three alignment targets at layer 16, including Q-Former outputs, intermediate projector features, and final projector outputs. As shown in Table~\ref{tab:ablation_align_target}, aligning with Q-Former outputs yields the best average accuracy of 66.08\%, outperforming intermediate projector features by 0.99 pp and final projector outputs by 1.17 pp. Q-Former outputs preserve richer geometric semantics from point clouds, while intermediate and final projector features are increasingly adapted toward the LLM input space, losing discriminative 3D structural information critical for downstream classification tasks.

\begin{table}[tb!]
\centering
\caption{Ablation study on the alignment target. We evaluate the impact of different alignment targets at layer 16 on classification accuracy (\%) using cosine similarity loss with $\lambda=0.1$.}
\vspace{-8pt}
\label{tab:ablation_align_target}
\resizebox{\linewidth}{!}{\begin{tabular}{lccccc}
\toprule
\multirow{2}{*}{Loss} & \multicolumn{2}{c}{ModelNet40} & \multicolumn{2}{c}{Objaverse} & \multirow{2}{*}{Average} \\
\cmidrule(lr){2-3} \cmidrule(lr){4-5}
& (I) & (C) & (I) & (C) & \\
\midrule
q-former & 61.55 & 60.78 & \textbf{72.50} & \textbf{69.50} & \textbf{66.08} \\
Projector-mid & \textbf{62.16} & \textbf{60.70} & 69.50 & 68.00 & 65.09 \\
Projector & 61.79 & 60.33 & 70.00 & 67.50 & 64.91 \\
\bottomrule
\end{tabular}}
\end{table}

\textbf{Feature quality analysis.} To further validate that alignment regularization improves the discriminative quality of point cloud tokens in LLM hidden layers, we conduct KNN classification experiments on ModelNet40. We extract point cloud tokens from layers 4, 8, 12, 16, 20, 24, 28, and 32 of both the baseline and aligned models, and evaluate their classification accuracy using K=1 and K=10 nearest neighbors. As shown in Figure~\ref{fig:knn}, the aligned model outperforms the baseline across all evaluated layers for both K=1 and K=10 scenarios.
For K=1, the aligned model achieves peak accuracy of 85.43\% at layer 20, compared to the baseline's 83.40\% at layer 20, demonstrating a 2.03 pp improvement. For K=10, the aligned model reaches 84.62\% at layer 12, while the baseline at 83.81\% at layer 12. Notably, the aligned model demonstrates sustained superior accuracy across the entire network depth, confirming that alignment regularization preserves richer discriminative geometric features throughout the LLM's forward pass, leading to better performance on the downstream tasks.

\textbf{Data efficiency.} To assess the effectiveness of our alignment 
regularization under varying data availability, we conduct experiments on 
the 3D object captioning task using different fractions of training data 
from the Objaverse dataset. As shown in Figure~\ref{fig:fraction}, the 
aligned model consistently outperforms the baseline across all data regimes. 
With only 10\% of training data, our method achieves 41.95\% compared to 
the baseline's 41.77\%, demonstrating effectiveness even in highly data-scarce 
scenarios. At 30\% data fraction, the aligned model gains 0.91 pp over 
the baseline. Most notably, with full training data, the aligned model achieves 53.05\% 
accuracy compared to the baseline's 48.17\%, representing a substantial 
4.88 pp improvement.

Critically, the baseline model exhibits a significant performance degradation from 52.72\% (50\% data) to 48.17\% (100\% data), losing 4.55 pp when provided with more training data. This suggests that the standard next-token prediction objective struggles to effectively utilize larger, more diverse datasets, potentially leading to optimization instability or overfitting. In stark contrast, our aligned model maintains consistent improvement across all data scales, reaching its peak performance at 100\% data. This strongly indicates that our alignment regularization acts as an effective regularizer, stabilizing the training process and enabling the model to fully leverage increased data availability, rather than simply overfitting to a larger dataset.


\vspace{-6pt}
\section{Conclusion}
\vspace{-4pt}
\label{sec:con}

In this work, we address the critical challenge of data scarcity in 3D vision-language models by proposing~\mname, a feature-level alignment regularization method. By introducing explicit supervision at intermediate layers through Q-Former output alignment, our method effectively prevents geometric information degradation with minimal computational overhead. Extensive experiments validate the effectiveness of our approach, achieving 2.08 pp improvement on average for classification tasks across ModelNet40 and Objaverse datasets, with notable gains of 7.50 pp on open-vocabulary classification and 4.88 pp on captioning evaluated by Qwen2-72B-Instruct. These results demonstrate that feature-level supervision enables more effective utilization of limited 3D data and enhances generalization across diverse 3D understanding tasks.


\textbf{Limitations. }The quality of alignment is constrained by the representational capacity of the visual feature aggregation module. Additionally, the alignment at a single fixed layer and the simple loss may not optimally supervise the entire feature transformation process. Future work includes exploring multi-layer alignment strategies, contrastive learning, and learnable distance metrics.


{
    \small
    \bibliographystyle{ieeenat_fullname}
    \bibliography{main}
}

\clearpage
\appendix
\section{Additional Experiments}

\subsection{Generalization to Non-Q-Former Architectures}
To validate that our alignment regularization is architecture-agnostic and not limited to Q-Former-based models, we apply our method to 3D-LLaVA~\cite{deng20253d}, which uses Omni Superpoint Transformer (OST) with MLP instead of Q-Former, on the Scan2Cap~\cite{chen2021scan2cap} dataset for scene-level 3D dense captioning. As shown in Table~\ref{tab:scene_level}, our method yields consistent improvements across all four metrics, with CIDEr@0.5 improving by 2.7, demonstrating effectiveness beyond object-level tasks and Q-Former architectures.

\begin{table}[h!]
\centering
\caption{Scene-level 3D dense captioning on Scan2Cap~\cite{chen2021scan2cap}. Our alignment regularization is applied to 3D-LLaVA~\cite{deng20253d}, which does not use Q-Former. (* denotes reproduced results.)}
\vspace{-8pt}
\label{tab:scene_level}
\begin{tabular}{lcccc}
\toprule
Method & C@0.5 & B-4@0.5 & M@0.5 & R@0.5 \\
\midrule
3D-LLaVA$^{*}$ & 76.1 & 36.3 & 27.0 & 57.3 \\
+ Ours & \textbf{78.8} & \textbf{37.1} & \textbf{27.2} & \textbf{57.6} \\
\bottomrule
\end{tabular}
\end{table}

\subsection{Generalization to Different LLM Backbones}
To verify that the middle-layer alignment strategy generalizes across LLM architectures, we conduct layer selection experiments on Phi-3 (32 layers). As shown in Table~\ref{tab:layer_phi3}, middle layers (12--20) generally outperform both early and late layers. This aligns with our Phi-2 ablation (Table~\ref{tab:ablation_layer}) where Layer~16 achieves optimal results, demonstrating that the strategy is robust across different LLM backbones.

\begin{table}[h!]
\centering
\caption{Layer selection ablation on Phi-3 backbone. Middle layers generally achieve better performance, aligning with the Phi-2 results in Table~\ref{tab:ablation_layer}.}
\vspace{-8pt}
\label{tab:layer_phi3}
\resizebox{\linewidth}{!}{
\begin{tabular}{lccccc}
\toprule
\multirow{2}{*}{Layer} & \multicolumn{2}{c}{ModelNet40} & \multicolumn{2}{c}{Objaverse} & \multirow{2}{*}{Average} \\
\cmidrule(lr){2-3} \cmidrule(lr){4-5}
& (I) & (C) & (I) & (C) & \\
\midrule
MiniGPT-3D & 62.50 & \textbf{56.50} & 57.13 & 48.03 & 56.04 \\
\midrule
8 & 65.00 & \textbf{56.50} & 57.18 & \textbf{49.59} & 57.07 \\
12 & \textbf{66.00} & 56.00 & \textbf{58.74} & 49.35 & \textbf{57.52} \\
16 & 65.50 & 56.00 & 58.69 & 48.66 & 57.21 \\
20 & \textbf{66.00} & 56.00 & 58.51 & 49.27 & 57.45 \\
24 & 65.00 & 55.00 & 57.82 & 49.31 & 56.78 \\
28 & 63.50 & 54.00 & 58.10 & 48.95 & 56.14 \\
\bottomrule
\end{tabular}
}
\end{table}

\end{document}